\documentclass{article} 
\usepackage{iclr2024_conference,times}

\usepackage{amsmath,amsfonts,bm}

\def\secref#1{section~\ref{#1}}

\def\eqref#1{equation~\ref{#1}}

\def\1{\bm{1}}

\DeclareMathAlphabet{\mathsfit}{\encodingdefault}{\sfdefault}{m}{sl}
\SetMathAlphabet{\mathsfit}{bold}{\encodingdefault}{\sfdefault}{bx}{n}

\usepackage{url}
\usepackage{multirow}
\usepackage{array}
\usepackage{graphicx}
\usepackage{enumitem}
\usepackage{amsmath,amsfonts,amssymb}
\usepackage{adjustbox}
\usepackage{microtype} 
\usepackage{lipsum} 

\usepackage{threeparttable}
\usepackage[citecolor=gray,filecolor=red, linkbordercolor=red,colorlinks=true]{hyperref}

\newcommand{\tabref}[1]{Table~\ref{#1}}
\newcommand{\figureref}[1]{Figure~\ref{#1}}
\newcommand{\tableref}[1]{Table~\ref{#1}}

\newcommand{\rthree}[1]{\textcolor{black}{#1}}
\newcommand{\rfive}[1]{\textcolor{black}{#1}}
\newcommand{\rsix}[1]{\textcolor{black}{#1}}
\newcommand{\reight}[1]{\textcolor{black}{#1}}
\newcommand{\sref}[1]{Section.~\ref{#1}}

\title{Long-Term Typhoon Trajectory Prediction: A Physics-Conditioned Approach Without Reanalysis Data}

\author{Young-Jae Park\thanks{Y.-J. Park and M.-S. Seo provided equal contributions to this work.}
\textsuperscript{*1}, 
\textbf{Minseok Seo}\textsuperscript{*2}, 
\textbf{Doyi Kim}\textsuperscript{2}, 
\textbf{Hyeri Kim}\textsuperscript{2}, 
\textbf{Sanghoon Choi}\textsuperscript{3}, 
\textbf{Beomkyu Choi}\textsuperscript{2}, 
\\\textbf{Jeongwon Ryu}\textsuperscript{2}, 
\textbf{Sohee Son}\textsuperscript{2}, 
\textbf{Hae-Gon Jeon}\textsuperscript{1}, 
\textbf{Yeji Choi}\textsuperscript{2}
\\\textsuperscript{1} Gwangju Institute of Science and Technology,
\textsuperscript{2} SI Analytics, 
\textsuperscript{3} The University of Manchester
 \\
}

\newcolumntype{M}[1]{>{\centering\arraybackslash}m{#1}}

\begin{document}

\maketitle
\begin{abstract}
In the face of escalating climate changes, typhoon intensities and their ensuing damage have surged.
Accurate trajectory prediction is crucial for effective damage control.
Traditional physics-based models, while comprehensive, are computationally intensive and rely heavily on the expertise of forecasters.
Contemporary data-driven methods often rely on reanalysis data, which can be considered to be the closest to the true representation of weather conditions.
However, reanalysis data is not produced in real-time and requires time for adjustment because prediction models are calibrated with observational data.
This reanalysis data, such as ERA5, falls short in challenging real-world situations. Optimal preparedness necessitates predictions at least 72 hours in advance, beyond the capabilities of standard physics models.
In response to these constraints, we present an approach that harnesses real-time Unified Model (UM) data, sidestepping the limitations of reanalysis data.
Our model provides predictions at 6-hour intervals for up to 72 hours in advance and outperforms both state-of-the-art data-driven methods and numerical weather prediction models.
In line with our efforts to mitigate adversities inflicted by \rthree{typhoons}, we release our preprocessed \textit{PHYSICS TRACK} dataset, which includes ERA5 reanalysis data, typhoon best-track, and UM forecast data.
%
%
%
\end{abstract}

\section{INTRODUCTION}
As global warming accelerates, the intensity of typhoons is on the rise~\citep{walsh2016tropical,ipcc2023}.
Accurate typhoon trajectory prediction is of paramount importance to allow enough time for emergency management and to organize evacuation efforts.
This warning period is more critical in countries that lack sufficient infrastructure to forecast typhoons.
Many nations utilize numerical weather prediction (NWP) models for typhoon trajectory forecasting and base their typhoon preparedness measures on this forecasting information.
NWP models make predictions for atmospheric variables like geopotential height, wind vectors, and temperature using atmospheric governing equations.
Forecasters employ the specific characteristics of typhoons in their NWP output to infer the typhoon trajectory. 

However, interpreting the outputs of NWP models depends on the expertise of the forecasters, as well as the use of other specialized tracking algorithms, such as the European Centre for Medium-Range Weather Forecasts (ECMWF) Tracker~\citep{bi2023accurate,lam2022graphcast}.
To address these issues, data-driven typhoon trajectory prediction methods have been proposed~\citep{huang2023mgtcf, ruttgers2019prediction}.
Additionally, methods to generate reanalysis data for applying ECMWF Tracker~\citep{ecmwf2021tropical} have also emerged~\citep{pathak2022fourcastnet, espeholt2022deep, lam2022graphcast, nguyen2023climax}.
Yet, they possess the following limitations:
Weather forecasting typically uses the ECMWF ReAnalysis-v5 (ERA5)~\citep{dee2011era} data for training and inference.
However, ERA5 data goes through post-correction processes, including data assimilation based on NWP forecast data. 
As a result, although ERA5 data shows high accuracy, it is not accessible until 3-5 days after the typhoon.

Consequently, they are unsuitable for real-time tasks like typhoon trajectory prediction.
In addition, existing data-driven real-time trajectory prediction models~\citep{ huang2022mmstn} restrict their forecasts to less than 24 hours.
This considerably shorter prediction window, compared to the NWP models, makes them incapable of planning effective damage control for typhoons.

%
%
%
%
%
%

\begin{figure*}[!t]
 \centering
 \includegraphics[width=0.9\linewidth]{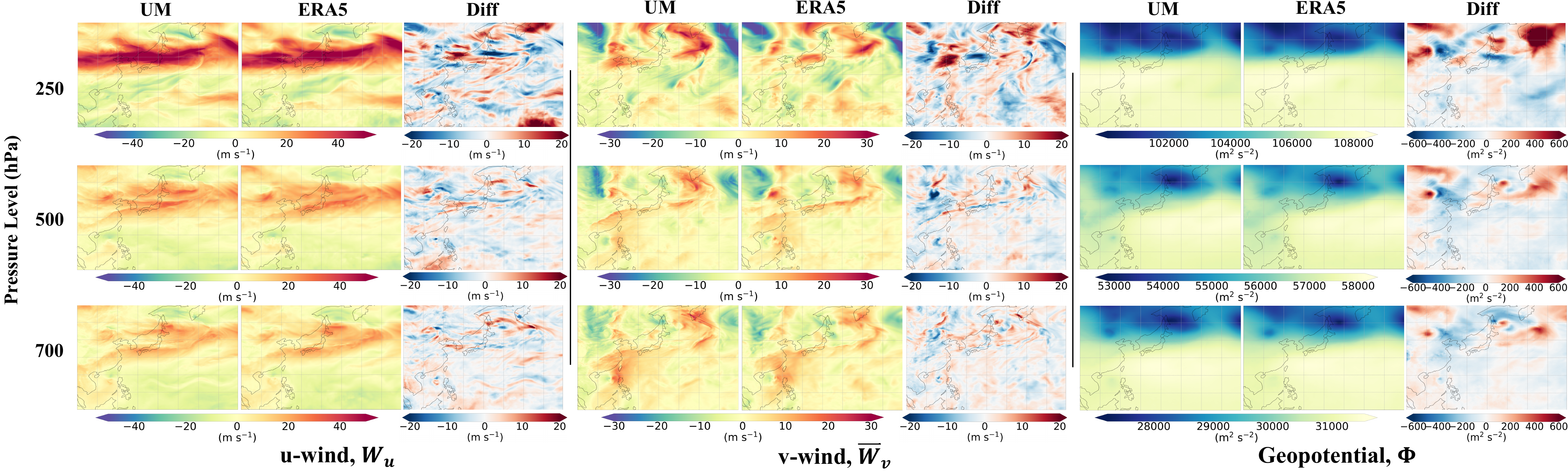}
 \caption{Visualization and comparison of the UM and ERA5 data at pressure levels of 250, 500, and 700 hPa, along with a difference map of two datasets. The UM forecast data is at a lead time of +72 hours\reight{,} and the ERA5 data corresponds to that forecasted time. \reight{This analysis covers the Western North Pacific basin, with latitudes ranging from 0 to 59.75°N and longitudes from 100°E to 179.75°E.}}
 \label{fig:fig1}
\end{figure*}

%
For data-driven typhoon trajectory prediction, the dataset known as \textit{Best Track}~\citep{knapp2010international} has been a cornerstone.
Documenting the intensity and central points of typhoons, its records extend back to 1950.
Correspondingly, the ERA5 reanalysis data, which provides invaluable insights as an additional physics-conditioned dataset, is accessible for the same time range.
In contrast, the \reight{Unified Model (UM)}~\citep{brown2012unified} dataset is produced in near-real time, including geopotential height, wind vectors, and other atmospheric variables using primitive equations such as the conservation of momentum, mass, energy, and water mass.
%
%
\reight{However, the UM dataset inherently has several limitations.}

: The UM data records only include data from 2010 to the present.
Furthermore, while the UM dataset adheres to physical formulas, it does not guarantee an exact representation of real-world values and can have potential errors~\citep{brown2012unified}.
\figureref{fig:fig1} and \tableref{tab:compare} compare the UM dataset with ERA5. As illustrated, the UM data significantly differs from ERA5 in several areas.
%
%
Meanwhile, as \tableref{tab:compare} reveals, while the UM data has only \rfive{a 3-hour data acquisition delay}, it exhibits larger errors.

%

%
%
In this study, we address these issues by introducing a Long-Term Typhoon Trajectory Prediction method (LT3P).
LT3P primarily consists of two components: a physics-conditioned encoder and a trajectory predictor.
The first component, the physics-conditioned encoder, encodes influential variables representing the typhoon's trajectory and intensity.
In this initial phase, we harness three atmospheric variables—geopotential height, zonal, and meridional wind—across three pressure levels (700, 500, and 250 hPa) sourced from the ERA5 dataset.
After this, all network parameters are frozen. The next step exclusively focuses on the bias corrector from UM data to ERA5.
This correction process is fine-tuned on UM's geopotential height and wind vector data from pressure levels 250, 500, and 700 — available since 2010 — combined with the Best Track dataset. 
The second component, the trajectory predictor, accepts the typhoon's central coordinates as input and forecasts its trajectories for the future 72 hours.
Features from the physics-conditioned encoder are cross-attended within this trajectory predictor, facilitating precise physics-conditioned typhoon trajectory predictions.
%
%

%
%

Three primary contributions can be summarized:
\begin{itemize}[leftmargin=*]

\item We propose, for the first time, a real-time +72 hours typhoon trajectory prediction model without reanalysis data.

\item We provide the preprocessed dataset \textit{PHYSICS TRACK}, and training, evaluation, and pretrained weights of LT3P.

\item In the +72 hours forecast, we achieved state-of-the-art results, outperforming the NWP-based typhoon trajectory forecasting models by significant margins.

\end{itemize}

\begin{table}[t!]
\caption{Comparison of ERA5 and NWP datasets: Highlighting the trade-off between real-time data availability and forecast accuracy. While ERA5 boasts high accuracy, it lacks immediacy. In contrast, NWP provides real-time data but may contain forecasting errors.}
\label{tab:compare}
\resizebox{\columnwidth}{!}{%
\begin{tabular}{c|ccccc}
\hline \hline
  & \textbf{Type} & \textbf{Data Availability} & \textbf{Real-time}  & \textbf{Limitations} 	 & \textbf{Forecast Accuracy}
  \\ \hline
 \textbf{ERA5} & Reanalysis Dataset & 1950 - Present &  No & Time delay for data availability   & $ \approx 95\% $ \\
 \textbf{UM} & NWP Model	 & Approx. 2010 - Present & Yes  &  Potential for model errors in forecasting   & Low  \\
 \textbf{IFS} & NWP Model    & 1974 - Present & Yes  &  Potential for model errors in forecasting   & High \\
 \textbf{GFS} & NWP Model	 & 2015 - Present & Yes  &  Potential for model errors in forecasting  & Middle \\
 \hline \hline
 \end{tabular}%
 }
 \end{table}

\section{RELATED WORK}
Typhoon trajectory prediction and human trajectory prediction are closely related in that they focus on modeling future trajectories.
Unlike the trajectories of general robots and vehicles~\citep{jiang2023motiondiffuser}, typhoon trajectories have high uncertainty because the direction and speed of movement are not deterministic~\citep{golchoubian2023pedestrian}.
These characteristics are similar to pedestrian trajectory prediction~\citep{gu2022stochastic} in that the direction and speed of pedestrian movement are not determined and its uncertainty is high.
However, while human predictions focus on capturing social interactions between agents, typhoon predictions consider overall atmospheric and geoscientific variables.
In this section, we explore a typhoon trajectory prediction that takes into account the benefits of two distinct fields.
\subsection{Trajectory Prediction}
Trajectory prediction has primarily evolved from a focus on human trajectory prediction~\citep{alahi2016social,gupta2018social, huang2019stgat, mangalam2020not, salzmann2020trajectron++, gu2022stochastic}, as it has been extensively applied to human-robot interaction systems~\citep{mavrogiannis2023core}.

SocialGAN~\citep{gupta2018social} employs a Generative Adversarial Network (GAN) framework for predicting realistic future paths. By leveraging latent space vectors, it is able to generate various outputs that capture the intricate nuances of socially acceptable and multimodal human behaviors.
PECNet~\citep{mangalam2020not} offers a distinct conditional VAE-based approach by transforming the latent spaces. It aims to strike a balance between the fidelity and diversity of predicted samples through its truncation trick in the latent space.
Trajectron++~\citep{salzmann2020trajectron++} provides a holistic methodology, predicting the probability of a latent distribution during inference. This robustness arises from its ability to adapt distribution parameters based on external input, such as prior trajectory data.
Graph Neural Network (GNN)-based approaches~\citep{huang2019stgat,mohamed2020social,shi2021sgcn} have attracted interest \rthree{in} modeling and predicting trajectories in scenarios with dense populations, such as pedestrian movements.
Recently, MID~\citep{gu2022stochastic} proposes a high-performing trajectory prediction method.
This method enhances the conventional trajectory prediction structure—which previously consisted only of \reight{long short-term memory (LSTM), multi-layer perceptron (MLP)}, and GNN—by integrating Transformer and diffusion processes.

%

However, when human trajectory prediction methods are applied to typhoon trajectory prediction, the results typically show poor performance~\citep{huang2022mmstn, huang2023mgtcf}.
Unlike pedestrian scenes with multiple interacting entities, typhoons usually appear singly or in pairs at most.
This difference implies that the detailed social modeling crucial for pedestrian predictions might be less important in the case of typhoons.
While GNN techniques excel in complex interactive settings, their efficacy is reduced by the inherent nature of typhoons.

\subsection{Typhoon Trajectory Prediction}
Forecasting the trajectory of typhoons is still a complex task, affected by many variables including weather conditions, sea surface temperatures, and topography~\citep{emanuel2007environmental}.
As studies have progressed, forecasting methods have improved, moving from basic statistical models to more advanced contemporary numerical approaches~\citep{wang2015statistical, chen2019development}.
Particularly, NWP systems, which leverage an abundance of meteorological data, have emerged as fundamental tools for typhoon trajectory prediction.
However, the interpretation of NWP model outputs often requires the expertise of meteorologists or relies on complex typhoon trajectory prediction algorithms.

On the other hand, data-driven approaches~\citep{pathak2022fourcastnet, espeholt2022deep, lam2022graphcast, nguyen2023climax} demonstrate notable performance in the fields of climate science, suggesting their potential applicability in forecasting typhoon trajectories.
Initial attempts integrate Recurrent Neural Networks (RNNs) with specific data types but often lead to unsatisfactory results~\citep{alemany2019predicting}.
Leveraging a broader range of meteorological data has enhanced prediction accuracy, even though this introduces challenges in data processing and encoding~\citep{giffard2020tropical}.
Notably, the recent leading approaches in typhoon trajectory prediction emphasize multi-trajectory forecasts, enabled through GANs~\citep{huang2022mmstn, dendorfer2021mg}
%
%

Despite these efforts, data-driven models still exhibit limitations, particularly with respect to their long-term forecast performance~\citep{chen2020machine, huang2023mgtcf} and real-time.
%
In this context, our work aims to pioneer long-term predictions by integrating physical values from UM data, which are outputs from a real-time-capable NWP model, with the power of data-driven methods.

\section{METHOD}
\rthree{The LT3P model takes the typhoon center coordinates \( C_{o} := \{ c\}_{i=1}^{t_{o}} = \{x_{\text{lon}}, y_{\text{lat}}\}_{i=1}^{t_{o}} \), where \( t_{o} \) represent the input time sequences, while \( x_{\text{lon}} \) and \( y_{\text{lat}} \) denote the longitude and latitude. Here, \( i \) represents the index in the time sequence.}
%
%
In LT3P, the inputs are the geopotential height and wind vector from the UM, represented as $ \Phi_P = \{ z_{p} \}_{p \in P, i=t_{\rthree{o}}+1}^{t_{\rthree{o}}+t_{f}}$ and 
$ \vec{W}_P = \{ w_{p} \}_{p \in P, i=t_{\rthree{o}}+1}^{t_{\rthree{o}}+t_{f}}$, respectively. Here, \( P \) refers to the pressure levels and we use it as \( P = \{250, 500, 700\} \text{hPa} \). $ t_{f} $ represent the output time sequences.
LT3P then outputs the future typhoon center coordinates  %
$ C_{f} := \{ u\}_{i=t_{\rthree{o}}+1}^{t_{\rthree{o}}+t_{f}} = \{x_{lon},y_{lat}\}_{i=t_{\rthree{o}}+1}^{t_{\rthree{o}}+t_{f}} $.

Hence, LT3P comprises a physics-conditioned model which extracts the representation from \( \Phi \) and \( \vec{W} \), and a trajectory predictor that forecasts the typhoon trajectory. LT3P is optimized through a dual-branch training strategy.
During a pre-trained phase, the physics-conditioned model is trained using ERA5 as in \citep{nguyen2023climax}.
Subsequently, in a bias correction phase, the trajectory predictor is trained to produce outputs akin to ERA5, even when UM data is used as the input.

\subsection{\rthree{Preliminaries}}
\label{sec:preliminarily}
The NWP forecasting system primarily hinges on a core prediction formula for the geopotential height \( \Phi \) and the wind vector \( \vec{W} \) based on \reight{conservation laws}:
\begin{equation}
\frac{\partial}{\partial t} \begin{bmatrix} \Phi, & \vec{W}, & \rho, & T, & q \end{bmatrix} = \mathcal{F} \left( \vec{W}, \Phi, \rho, T, q \right),
\end{equation}
where \( \mathcal{F} \) refers to interactions among variables following the conservation laws.
The variables \( \rho, T, \) and \( q \) denote air density, temperature, and specific humidity, respectively.

Conservation equations underpin the NWP's forecasting. For instance, a momentum conservation with the spatial gradient \( \nabla \) is expressed by:
\begin{equation}
    \frac{\partial \vec{W}}{\partial t} = -\vec{W} \cdot \nabla \vec{W}\frac{\nabla P}{\rho} + \vec{g},
\end{equation}
and a mass conservation is below:
\begin{equation}
    \frac{\partial \rho}{\partial t} + \nabla \cdot (\rho \vec{W}) = 0.
\end{equation}
Other conserved quantities include energy, related to geopotential height \( \Phi \) and temperature \( T \), and water vapor content or specific humidity \( q \).

\textbf{Note on Accuracy and Time $t$.}
In the NWP system, a time \( t \) indicates progression. For longer predictions up to +72 hours, their accuracy depends on various factors: initial conditions, physics parameterization schemes, spatial resolution, and so on.
While the conservation law captures large-scale atmospheric dynamics effectively, it might not work on small-scale details due to local conditions like microclimates, terrain variations, or transient phenomena.

The Earth's atmosphere is a chaotic system~\citep{lorenz1963deterministic}.
%
Despite relying on conservation laws, NWP models are inevitably prone to errors due to the inherent imperfections in our understanding of physical processes and the true state of the atmosphere.
These discrepancies underscore the importance of data-driven approaches, in spite of the potential error in the UM data.
%
%

\subsection{Pre-Training on ERA5 Dataset}
\label{sec:Pre-Training}
\begin{figure*}[!t]
 \centering
 \includegraphics[width=0.7\linewidth]{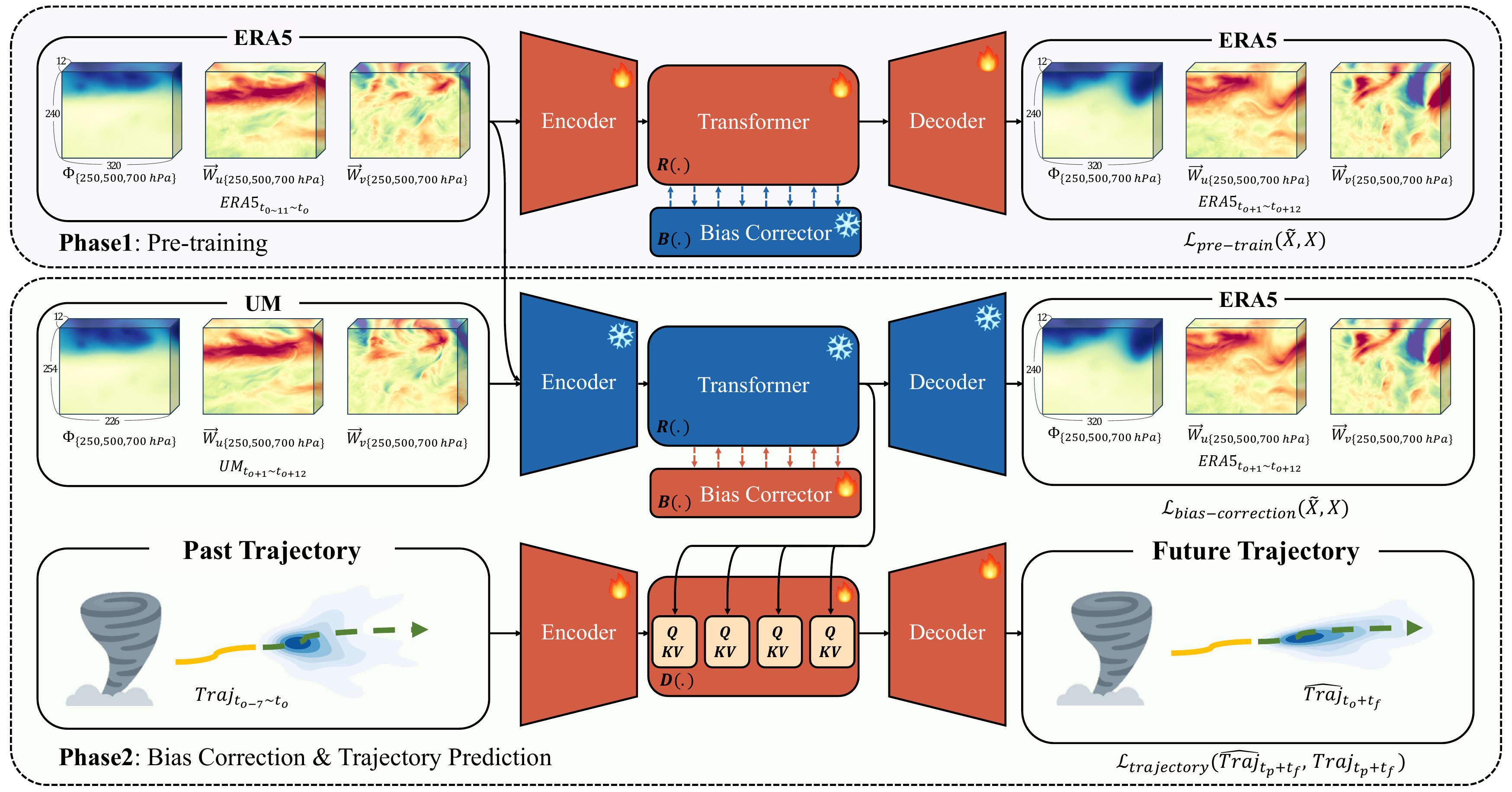}
 \caption{Overview of LT3P. In phase 1, the physics-conditioned model \( R(\cdot)\) is trained for the weather forecasting task to encode information from \(\Phi\), \( \vec{W}_{u}\), and \( \vec{W}_{v}\). Subsequently, in phase 2, using UM as the input, the typhoon trajectory prediction model \( D(\cdot)\) is trained in conjunction with the bias-corrector \( B(\cdot)\) to make accurate predictions.
}
 \label{fig:sample2}
\end{figure*}
\rthree{The ERA5 dataset, accumulated at 6-hour intervals from 1950 to the present, requires a significant amount of time to train a model from scratch.
%
%
%
%
To address this challenge, we have adopted a strategy of constructing a foundational model for typhoon trajectory prediction}.

%
%
%

%
The physics-conditioned model of LT3P is trained for a weather forecasting task, similar to~\citep{nguyen2023climax}, to effectively learn the representations of the highly-related variables for typhoon trajectory prediction in~\citep{giffard2020tropical}: $\Phi$, $ \vec{W}_{u}$, and $ \vec{W}_{v}$.

As shown in~\figureref{fig:sample2}, the physics-conditioned model consists of an encoder-transformer-decoder architecture. The encoder and decoder are built using 3D convolutions for spatio-temporal modeling, and the transformer is also constructed as a 3D transformer.
%
%

The objective of the physics-conditioned model is as follows:
\begin{equation}
    L_{pre\_train} = \frac{1}{T \times V \times H \times W} \sum_{i=1}^{T} \sum_{j=1}^{V} \sum_{k=1}^{H} \sum_{l=1}^{W} (\tilde{X}_{t_{f}}^{i,j,k,l}- X_{t_{f}}^{i,j,k,l})^{2},
\end{equation}
where \( T \) is the final lead time, \( V \) represents 9 variables, with \(\Phi, \vec{W}_{u}\), and \(\vec{W}_{v}\) for each of the 3 pressure levels. \( H \) and \( W \) denote height and width, respectively. 
\( \tilde{X} \) is the forecasting value for ERA5 from time \( t \) to \( t + t_{f} \) in the physics-conditioned model, given \( V \) as input from the period \( t - t_{f} \) to \( t \). \( X \) is its ground truth (GT).

\subsection{Bias correction \& Typhoon Trajectory Prediction}
\textbf{Bias correction.} Since ERA5 is reanalysis data, this limits the applicability on real-time tasks, which makes us use the UM data.
Hence, the typhoon trajectory prediction model, LT3P, accepts two primary inputs: the representation encoded by the 
\( R(\cdot) \) ---derived from the physics formula presented in \sref{sec:preliminarily} using \( \Phi \) and \( \vec{W} \)---and the typhoon's center coordinates \( C_{o} \). The model then produces the output \( C_{f} \).

However, since our \( R(\cdot) \) was trained with the ERA5 data, it may struggle with extracting meaningful representations from the UM data due to their domain gap.
To solve this problem, LT3P uses the bias-corrector \( B(\cdot) \) to adapt the variables generated through UM to the variables of ERA5.
The \( B(\cdot) \) has the following objective functions:
\begin{equation}
    L_{Bias\_correction} = \frac{1}{T \times V \times H \times W} \sum_{i=1}^{T} \sum_{j=1}^{V} \sum_{k=1}^{H} \sum_{l=1}^{W} (\hat{X}_{t_{f}}^{i,j,k,l}- X_{t_{f}}^{i,j,k,l})^{2},
\end{equation}
where \(\hat{X}\) is the ERA5 prediction which is bias-corrected version of the UM dataset.
%
%

\textbf{Typhoon Trajectory Prediction.}
In the second phase, LT3P takes ERA5 data from 1950 to 2010 and UM data from 2010 to 2018 as inputs in order to perform typhoon trajectory prediction.
The main model of LT3P, which predicts the center coordinates of the typhoon, generates the final trajectory by applying a cross-attention between the 3D features from the \( R(\cdot) \)'s transformer and the 1D features of the typhoon coordinates from the LT3P model's encoder.
The cross attention enables LT3P to take fully advantage of both data-driven and physics-based prediction. It also allows us to fuse these two features regardless of the shape or length of the features. The cross attention is defined as:
\[
\text{Attention}(Q, K, V) = \text{softmax} \left( \frac{Q K^T}{\sqrt{d}} \right) \cdot V,
\]
where \( Q = W^{(i)}_Q \cdot \phi_i(R(\Phi, \vec{W})) \), \( K = W^{(i)}_K \cdot \tau_\theta(C_{p}) \), and \( V = W^{(i)}_V \cdot \tau_\theta(C_{p}) \).  The \( \theta\) represents the set of parameters associated with the \(D (\cdot) \).
Here, \( \phi_i(z_t) \in \mathbb{R}^{N \times d_i} \) denotes a (flattened) intermediate representation of the transformer implementing \( \theta \). \( W^{(i)}_V \in \mathbb{R}^{d \times d_i} \), \( W^{(i)}_Q \in \mathbb{R}^{d \times d_\tau} \), and \( W^{(i)}_K \in \mathbb{R}^{d \times d_\tau} \) are learnable projection matrices.

Therefore, in LT3P, the final objective function is defined as follows:
\begin{equation}
    L_{total} = L_{Bias\_correction} + L_{trajectory}.
\end{equation}
Note that various trajectory prediction baselines such as GAN, CVAE, and diffusion-based models can be applied, whose effectiveness will be validated in \secref{sec:ablation}.
%

%

%
\begin{table}[t!]
\centering
\caption{Ensemble typhoon trajectory prediction results (FDE): Note that while the original MGTCF study only includes 24-hour predictions, for comparison purposes, we modified it to input 48 hours at 6-hour intervals and predict +72 hours at 6-hour intervals. \textbf{Bold} indicates the best performance, and \underline{underline} indicates the second-best performance. Note that \textbf{Dataset} refers to the dataset used for inference.}
\label{tab:main}
 \begin{threeparttable}
\resizebox{1.0\columnwidth}{!}{%
\begin{tabular}{c|c|ccc|cccccccccccc}
\hline\hline
\multirow{2}{*}{\textbf{Method}} & \multirow{2}{*}{\textbf{Real-time}} &\multicolumn{3}{c|}{\textbf{Dataset}} & \multicolumn{12}{c}{\textbf{Distance (km)}}   \\ 
&  & BST & ERA5 & UM & 6 h & 12 h & 18 h & 24 h & 30 h & 36 h & 42 h & 48 h & 54 h & 60 h & 66 h & 72 h \\ \hline 
SocialGAN~\citep{gupta2018social}-Ens &$\checkmark$ & $\checkmark$& - & -& 33.16 & 63.25 & 93.33 & 131.31 & 176.71 & 226.24 & 288.78 & 353.79 & 431.03 & 519.93 & 611.67 & 716.90 \\ 
STGAT~\citep{huang2019stgat}-Ens     & $\checkmark$& $\checkmark$& - &- & 56.68 & 100.51 & 160.08 & 209.83 & 269.81 & 325.44 & 377.86 & 435.69 & 492.44 & 550.37 & 613.97 & 682.16 \\ 
PECNet~\citep{mangalam2020not}-Ens   & $\checkmark$ & $\checkmark$ & -&- &41.61  &  56.74 &  82.26 & 112.28& 147.96 & 185.92 & 232.79 & 279.29 & 330.84 & 388.25 & 445.61 & 518.46 \\ 
MID~\citep{gu2022stochastic}-Ens     & $\checkmark$ & $\checkmark$ &- &- &109.31  &  173.35 &  230.45 & 289.62 & 359.10 & 422.11 & 490.93 & 564.23 & 638.04 & 712.97 & 798.63 & 881.34 \\ 
MMSTN~\citep{huang2022mmstn}-Ens     &$\checkmark$ & $\checkmark$ & -& -& 49.72
 & 109.23 &  173.64 & 246.26 & 334.62 & 430.45 & 552.15 & 667.34 & 819.22 & 969.13 & 1126.26 &   1300.59      \\ 
MGTCF~\citep{huang2023mgtcf}-Ens  &  $\times$   & $\checkmark$ & $\checkmark$&- & 46.12& 81.68 & 116.81 & 155.72 & 201.26 & 249.42 & 302.19 & 361.14 & 427.35 & 499.85 & 576.80 & 655.48 \\ \hline

JTWC~\citep{chen2023evaluation}    &  $\checkmark$   & -& - &- & -  &   -   &   -   &87.09&   -   &   -   &   -   &146.45&   -   &  -    &   -   &217.05\\

\hline
JMA-GEPS~\citep{chen2023evaluation}  & $\checkmark$ &- & - & -&-   &   62.83   &- & 98.10 &   -   &  138.26    &   -   & 180.04 &   -   &  222.75    &  -    & 259.07\\ 
ECMWF-EPS~\citep{chen2023evaluation}      & $\checkmark$  & - & - & - & - &   57.75   & - & 78.87 &  -    &  108.28    &   -   & 138.97 &    -  &  176.02    &   -   & 210.71\\ 
NCEP-GEFS~\citep{chen2023evaluation}    & $\checkmark$  &- & - &- & -  &   52.34   & - & 76.17 &    -  &  105.55    &  -    & 142.26 &    -  &  185.36    &  -    & 225.59\\
UKMO-EPS~\citep{chen2023evaluation}    & $\checkmark$  & -& -& -&  - &   62.74   & - & 90.90 & -     &  122.90    &  -    & 160.41 &   -   &  200.89    &    -  & 245.28\\
\hline
LT3P (UM Only)-Ens& $\checkmark$ & $\checkmark$ & - & $\checkmark$& 42.20 &   58.83   &  88.39    & 114.84     & 147.39     &  189.03    &  220.41   &  255.57    &  320.47    & 334.88     &  370.64    &  390.92    \\
LT3P (Bias-corrected UM)-Ens & $\checkmark$ & $\checkmark$ & - & $\checkmark$ & \underline{6.30}  &  \underline{19.45}    &  \underline{22.83}   &   \underline{39.81}   &   \underline{50.29}  &   \underline{60.76}   &  \underline{65.38}    &   \underline{70.01}   &   \underline{83.98}   &   \underline{100.88}   &   \underline{110.36}   &  \underline{143.03}    \\ 
LT3P (ERA5 Only)-Ens    & $\times$ & $\checkmark$ & $\checkmark$& -&\textbf{5.29}   &   \textbf{6.80}   &   \textbf{12.34}  &  \textbf{19.90}    &  \textbf{24.09}    &  \textbf{30.42}    &   \textbf{35.92}   &  \textbf{39.29}    &  \textbf{44.83}    &  \textbf{50.30}    &  \textbf{60.11}    & \textbf{70.94}      \\
\hline \hline
\end{tabular}%
}
    \begin{tablenotes}
    \item[*] \scriptsize Note that best track data from MMSTN and MGTCF could not be obtained because they use data from specific organizations. \\ Therefore, the experimental results of each of the two studies were evaluated only in 2019.
    \end{tablenotes}
 \end{threeparttable}
\end{table}
\section{Evaluations}
In this section, we first describe the dataset and the pre-processing methods used as well as an implementation detail of our LT3P.
We then show quantitative and qualitative evaluations and provide an extensive ablation study to check the effectiveness of each component in LT3P.
\subsection{Experimental Settings}
\textbf{\textit{Physics Track} dataset.} 
We collect the center coordinates of typhoons at 6-hour intervals (00, 06, 12, 18) from 1950 to 2021.
We use data corresponding to the Western North Pacific basin, with latitudes ranging from 0 to 59.75°N and longitudes from 100°E to 179.75°E, \rsix{as typhoons overwhelmingly occur within this region.~\citep{stull2011meteorology}}
Any typhoon that dissipated within 48 hours is excluded from the dataset.
\reight{The training data uses years 1950-2018 and comprises 1,334 typhoons, while the test dataset covers years 2019-2021 and includes 90 typhoons.
Note that for hyperparameter tuning, we train on data from 1950 to 2016 and set the 2017 to 2018 data as the validation set.
Subsequently, we evaluate the final performance of model using the entire dataset from 1950 to 2018.}

In sync with the duration of the typhoon, we utilize variables in the ERA5 dataset which are closely associated with typhoons, notably the geopotential height and wind vectors.
Data for the pressure levels at 250, 500, and 700 hPa are collected.
For training purposes, the ERA5 dataset is employed from 1950 to 2018, and the data spanning 2019, 2020, and 2021 is set aside.
Regarding the real-time data from the UM, we compile data from the years 2010 to 2021, ensuring the inclusion of the same variables and pressure levels as found in the ERA5 dataset.

\textbf{\textit{Note:}} All variables in ERA5 and UM are normalized using the formula \((\text{variable} - \text{mean})/\text{std}\) because no specific minimum-maximum range is predetermined for each variable. 

For both the ERA5 and UM datasets, information is extracted to match the conditions of the best track.
Therefore, the input dimension for ERA5 and UM is set to $(B \times T \times V \times H \times W) \rightarrow (B \times 12 \times 9 \times 240 \times 320)$.
\rthree{Note that the resolution has been adjusted using bi-linear interpolation.}
%

\textbf{Evaluation metrics.} 
We compare our LT3P with global meteorological agencies, including the Joint Typhoon Warning Center (JTWC), Japan Meteorological Agency (JMA), European Centre for Medium-Range Weather Forecasts (ECMWF), National Centers for Environmental Prediction (NCEP), and the United Kingdom Meteorological Office (UKMO).
Additionally, the Global Ensemble Forecast System (GEPS) refers to the results of 20 separate generated forecasts~\citep{zhou2022development} and then combines them into an ensemble \rthree{average} forecast, which is to average the sampled forecasting results.
Therefore, the data-driven models are evaluated using the ensemble \rthree{average} method to benchmark their performance compared to the ensemble \rthree{average} NWP models.
%
%
%
Performance is measured at 6-hour intervals, consistent with existing meteorological agencies.
%

We use (1) Average Displacement Error (ADE) - average Euclidean distance between a prediction and ground-truth trajectory; (2) Final Displacement Error (FDE) - Euclidean distance between a prediction and ground-truth final destination.
%
%

\begin{table}[t!]
\centering
\caption{Results of Stochastic Typhoon Trajectory Predictions (FDE): This table presents results generated from 20 samples, consistent with the conventional approach in NWP-based GEPS~\citep{zhou2022development}, which employs 20 ensemble members for the final prediction.}
\label{tab:sto}
\resizebox{0.8\columnwidth}{!}{%
\begin{tabular}{ccccccccccccc}
\hline\hline
\multirow{2}{*}{\textbf{Method}} & \multicolumn{12}{c}{\textbf{Distance (km)}}                                               \\ 
                        & 6 h & 12 h & 18 h & 24 h & 30 h & 36 h & 42 h & 48 h & 54 h & 60 h & 66 h & 72 h \\ \hline 
SocialGAN~\citep{gupta2018social} & 17.59
 & 30.30 & 40.97 & 50.30 & 64.40 & 71.63 & 88.00 & 97.81 & 110.49 & 126.92 & 144.66 & 164.99 \\ 
STGAT~\citep{huang2019stgat}    & 32.81 & 55.24 & 86.86 & 110.93 & 147.86 & 180.03 & 204.79 & 233.53 & 259.93 & 277.09 & 298.62 & 322.54 \\ 
PECNet~\citep{mangalam2020not} &  28.40  &  34.34 &  45.88 & 57.36 & 69.44 & 81.17 & 97.84 & 112.34 & 130.22 & 150.18 & 169.22 & 188.74 \\ 
MID~\citep{gu2022stochastic} &  24.54
  &  42.66 &  56.52 & 73.41 & 96.04 & 116.86 & 136.44 & 160.55 & 184.09 & 205.78 & 230.09 & 256.26 \\ \hline
MMSTN~\citep{huang2022mmstn}& 19.46 & 42.44 & 67.38 & 93.59 &  128.97 & 162.90 &  193.29 &  232.58 &  269.17 &  309.55 &   348.10 & 384.57 \\ 
MGTCF~\citep{huang2023mgtcf} & 20.84&  37.59 & 52.45 & 67.29 & 83.09 & 104.13 & 128.90 & 156.98 & 186.62 & 218.18 & 249.22 & 280.90 \\ \hline
LT3P (Bias-corrected UM) &  \underline{1.97}                 & \underline{2.31}    &  \underline{3.32}    &  \underline{8.80}    &  \textbf{17.89}    & \textbf{25.53}     &   \underline{30.57}   &  \textbf{36.39}   &  \underline{41.12}    &   \textbf{47.38}   &  \textbf{51.87}         & \textbf{65.24}     \\
LT3P (ERA5 Only)     &       \textbf{0.66} & \textbf{1.12} & \textbf{2.26} & \textbf{2.34} & \underline{21.93}& \underline{26.36}& \textbf{26.99}& \underline{41.39}& \textbf{40.24}& \underline{49.63}& \underline{52.97}& \underline{65.95}

      \\ \hline \hline
\end{tabular}%
}
    \begin{tablenotes}
    \item[*] \scriptsize * Note that best track data from MMSTN and MGTCF could not be obtained because they use data from specific organizations. \\ Therefore, the experimental results of each of the two studies were evaluated only in 2019.
    \end{tablenotes}
    \vspace{-8mm}
\end{table}

\textbf{Implementation details.}
The physics-conditioned model \( R(\cdot)\) is comprised of an encoder, translator and decoder.
The encoder and decoder are composed of {\fontfamily{qcr}\selectfont(Conv, LayerNorm, SiLU)}~\citep{elfwing2018sigmoid}) and {\fontfamily{qcr}\selectfont(Conv, LayerNorm, SiLU, PixelShuffle)}, respectively.
In the experiments, the encoder and decoder consist of four blocks each.
The translator is selected as the transformer block and has a total of three blocks.
The trajectory predictor \(D(\cdot) \) follows GAN~\citep{gupta2018social}, CVAE~\citep{mangalam2020not} and diffusion~\citep{gu2022stochastic}, respectively, and the experimental results in~\tabref{tab:main} and~\tabref{tab:sto} are obtained from the diffusion-based predictor.
%

For training, we use a machine with 8 Nvidia-Quadro RTX 8000 GPUs.
The batch size is set to 128, and we use Adam optimizer with a learning rate of 0.001.
Additionally, an Exponential Moving Average with a momentum of 0.999 is applied, and the training is conducted for 2,000 epochs.
Note that we did not employ data augmentation, as the geometric and intensity values in typhoon forecasting are extremely sensitive.
Applying data augmentation might compromise the integrity of physical rules.
\reight{For fair comparison, the data-driven baseline models such as SocialGAN~\citep{gupta2018social}, STGAT~\citep{huang2019stgat}, PECNet~\citep{mangalam2020not}, MID~\citep{gu2022stochastic}, and MMSTN~\citep{huang2022mmstn} are trained and evaluated on the best track dataset, while MGTCF~\citep{huang2023mgtcf} utilizes both the best track and ERA5 datasets.}

\subsection{Experimental Results}
\paragraph{Quantitative results.} 
In~\tabref{tab:main}, we compare the performance of LT3P with the state-of-the-art human trajectory prediction methods as well as data-driven typhoon prediction methods.
The table also includes forecast results from several renowned meteorological agencies, including JTWC, JMA, ECMWF, NCEP, GEFS, and UKMO, which are based on the conservation laws in ~\sref{sec:Pre-Training}.
Our observations indicate that LT3P (ERA5 Only) exhibits the most promising results, with LT3P (Bias-corrected UM) trailing closely behind.
It is important to highlight that MID, which has achieved the best results among data-driven human trajectory predictors, performs poorly in ensemble forecasting.
This discrepancy arises from the inherent goal of human trajectory prediction models.
They have developed to capture social relations between agents, and its probabilistic distribution attempts to cover a wide and diverse distribution to represent all feasible socially-acceptable paths and destination.

Furthermore, it is noteworthy that our model consistently outperforms the numerical prediction models of the established operational forecast centers.
These experimental results demonstrate the potential of real-time NWP data to serve as an alternative to reanalysis data in the field of real-time typhoon trajectory prediction.
Note that studies in~\citep{bi2023accurate, lam2022graphcast}, which predict reanalysis data and then use the ECMWF-tracker for typhoon trajectory prediction, are excluded in our comparisons because code and weights are unavailable, and there are no access to the ECMWF-tracker~\citep{ecmwf2021tropical}.

\tabref{tab:sto} shows the performance results from our stochastic trajectory prediction and the existing approaches to data-driven trajectory prediction. We report the results have the lowest error among 20 generated trajectories. 
As revealed in~\tabref{tab:sto}, LT3P stands out as the state-of-the-art model for stochastic trajectory prediction.
We also observe significant performance improvements in other data-driven trajectory prediction models.
These encouraging results can be attributed to the inherent strength of existing data-driven methods, which are designed to generate a diverse and wide range of trajectory predictions.

\begin{figure*}[t!]
 \centering
 \includegraphics[width=0.8\linewidth]{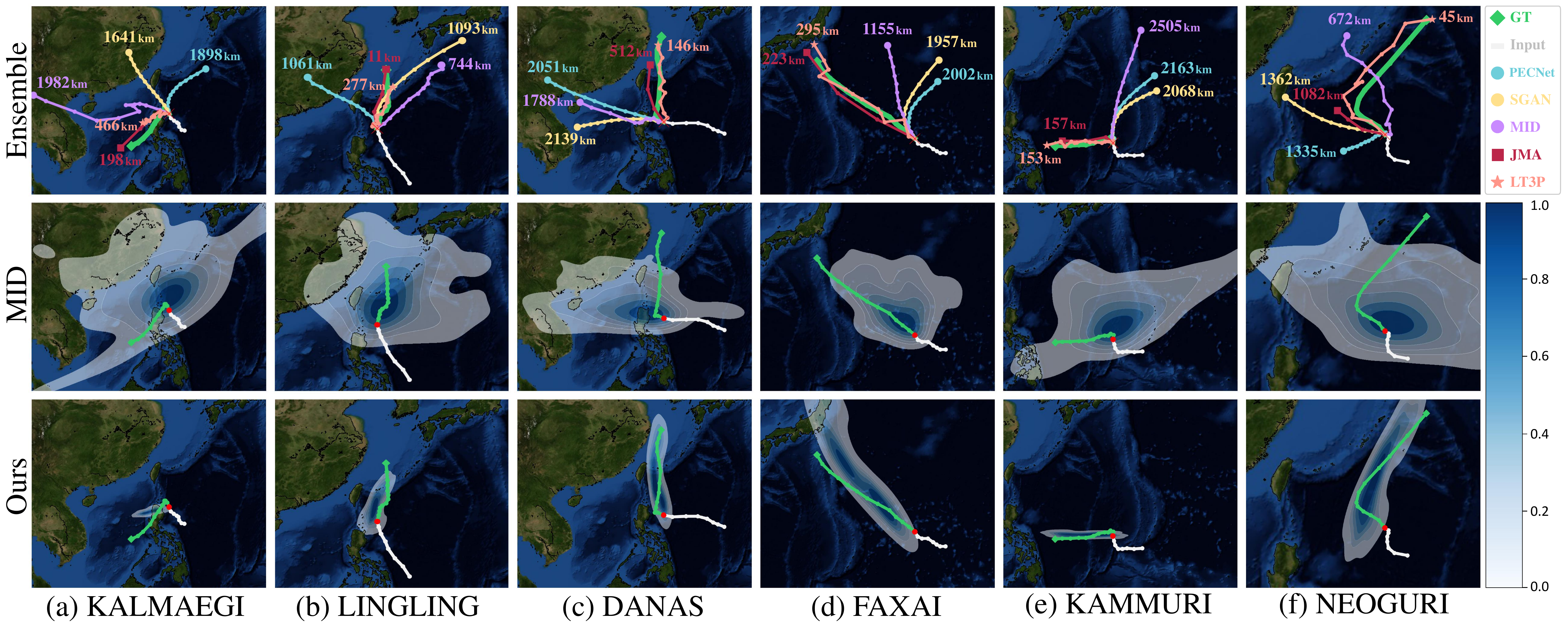}
 \caption{Qualitative analysis between LT3P and other trajectory prediction baselines. Note that the typhoon trajectory provided by JMA exists at 24-hour intervals, so the three coordinates have been linearly connected. Additionally, the error (in km) of the final coordinate is also indicated. \reight{The probability map is visualized using kernel density estimation (KDE).}}
 \label{fig:vis4}
\end{figure*}

\paragraph{Qualitative Results}
\figureref{fig:vis4} presents the qualitative results for ensemble and stochastic predictions.
In~\figureref{fig:vis4}-(c), only the JMA model and our LT3P predict the change in trajectory of typhoons like DANAS well.
These results highlight the importance of physics values in typhoon trajectory prediction.
In~\figureref{fig:vis4}-(f), LT3P superiority is also confirmed as well.
In addition, MID, using only coordinates, outperforms the physics-based JMA.
Such instances show the promising potential of data-driven models.
Additionally, in the case of stochastic results, LT3P does not aim for a wide and diverse range of predictions, but focuses on generating forecasts that closely approximate the actual typhoon trajectory.
These qualitative results are supporting evidence for LT3P's strong performance in ensemble forecasting, as reported in~\tabref{tab:main}.

\begin{figure*}[t!]
 \centering
 \includegraphics[width=1.0\linewidth]{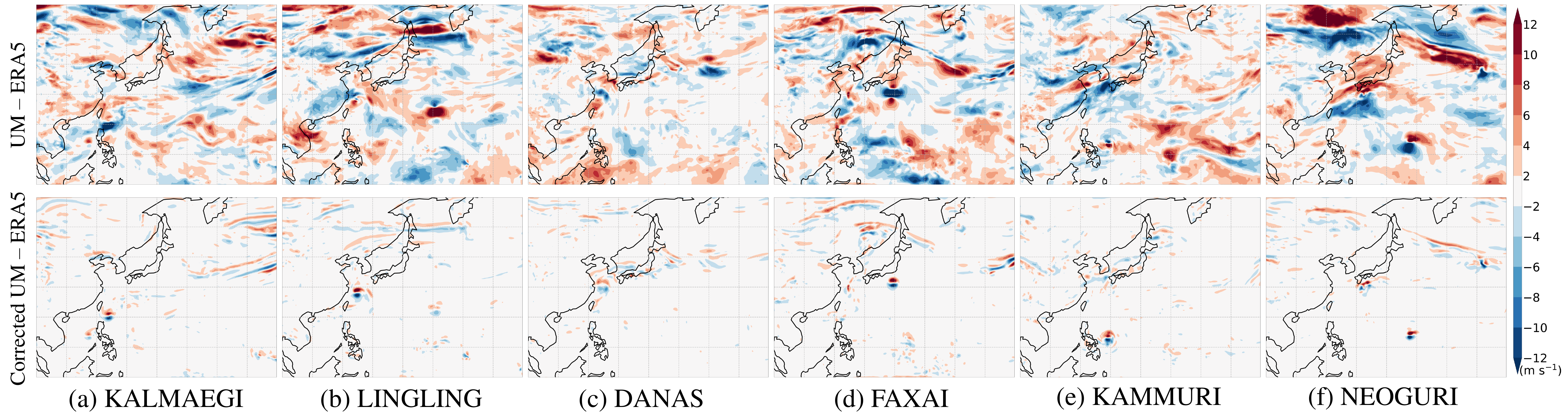}
 \caption{\reight{Zonal wind bias of the UM before and after bias correction with respect to ERA5.}}
 \label{fig:vis5}
\end{figure*}

~\figureref{fig:vis5} presents the differences in zonal wind field between UM 72-hour forecast and ERA5 before and after the bias correction of UM.
\reight{The figure shows that the UM exhibits a significantly smaller bias after the bias-correction phase than the original UM field.}
%
These show that the atmospheric conditions across the entire region, including the trajectories of the typhoons, have been well corrected in LT3P.
%

%
%

\subsection{Ablation Study}
\label{sec:ablation}
To assess the efficacy of each component within the LT3P framework, a systematic analysis is conducted: a baseline trained solely on the UM, a joint training scheme trained on ERA5 from 1950-2009 and the UM from 2010-2018, an extended method incorporating phase 1 pre-training and, the full LT3P pipeline with the bias correction.
The results are reported in~\tabref{tab:able}.
The empirical findings in~\tabref{tab:able} suggest that all components, barring the UM Only training, yield good results.
In particular, the LT3P configuration, which leverages all the components, achieves the best performance.

\tabref{tab:model} presents the experimental results when applying well-known trajectory prediction models such as GAN, CVAE, and diffusion-based methods to the LT3P framework.
In~\tabref{tab:model}, the diffusion-based method achieves the better performance than the others.
%
Moreover, all the methods, including GAN, CVAE, and diffusion, exhibit significant performance improvements when incorporating UM data compared to using only coordinate data (refer to the prediction result on 72 hours in~\tabref{tab:main}).
These findings indicate that the performance gains can be attributed to the LT3P framework itself rather than to the backbone models.
%

\begin{table}[t!]
  \centering
  \begin{minipage}{0.55\linewidth}
    \centering
    \caption{Ablation study on the efficiency of each component of LT3P.}
    \label{tab:able}
    \resizebox{\columnwidth}{!}{%
    \begin{tabular}{cccc|c}
    \hline \hline
    \multicolumn{4}{c|}{\textbf{Component}} & \textbf{Metric}  \\ \hline
    UM & Joint Training & Pre-Training & Bias-correction & ADE/FDE (km) \\ \hline
    \checkmark & & & & 219.46 / 390.92 \\
    \checkmark & \checkmark & & & 80.39 / 190.75 \\
    \checkmark & \checkmark & \checkmark & & 85.62 / 198.11 \\
    \checkmark & \checkmark & \checkmark & \checkmark & \textbf{65.25} / \textbf{143.03} \\ \hline \hline
    \end{tabular}%
    }
  \end{minipage}%
\hspace{8mm}
  \begin{minipage}{0.35\linewidth}
    \centering
    \caption{Ablation study on backbone models.}
    \label{tab:model}
    \resizebox{\columnwidth}{!}{%
    \begin{tabular}{c|c}
    \hline \hline
    \multirow{2}{*}{\textbf{Backbone}} & \textbf{Metric}  \\ \cline{2-2}
    & ADE/FDE (km) \\ \hline
    SocialGAN~\citep{gupta2018social} & 69.99 / 155.23 \\
    PECNet~\citep{mangalam2020not} & 69.81 / 152.07 \\
    MID~\citep{gu2022stochastic} & \textbf{65.25} / \textbf{143.03} \\ \hline \hline
    \end{tabular}%
    }
  \end{minipage} 
\end{table}

\section{Conclusion}
We propose the Long-Term Typhoon Trajectory Prediction (LT3P) for real-time typhoon trajectory prediction without reanalysis data.
To the best of our knowledge, LT3P is the first data-driven typhoon trajectory prediction that employs a real-time NWP dataset.
Unlike methods that generate reanalysis data, LT3P predicts the typhoon's central coordinates without any need for an additional forecaster and algorithms.
In addition, our LT3P does not require weather forecasting experts, making it easily accessible for various institutions with limited meteorological infrastructure.
Using extensive evaluations, we confirm that our LT3P achieved state-of-the-art performance in typhoon trajectory prediction.
However, our model has only been applied to typhoons and has not been applied to other regions of tropical cyclones. In future work, we plan to apply it to all kinds of tropical cyclones.
Lastly, we hope to contribute to the field of climate AI by releasing our \textit{PHYSICS TRACK} dataset, training, test codes of LT3P, and pretrained weights to the public.
\bibliography{iclr2024_conference}

\begin{thebibliography}{37}
\providecommand{\natexlab}[1]{#1}
\providecommand{\url}[1]{\texttt{#1}}
\expandafter\ifx\csname urlstyle\endcsname\relax
  \providecommand{\doi}[1]{doi: #1}\else
  \providecommand{\doi}{doi: \begingroup \urlstyle{rm}\Url}\fi

\bibitem[Alahi et~al.(2016)Alahi, Goel, Ramanathan, Robicquet, Fei-Fei, and Savarese]{alahi2016social}
Alexandre Alahi, Kratarth Goel, Vignesh Ramanathan, Alexandre Robicquet, Li~Fei-Fei, and Silvio Savarese.
\newblock Social lstm: Human trajectory prediction in crowded spaces.
\newblock In \emph{Proceedings of the IEEE/CVF conference on computer vision and pattern recognition}, 2016.

\bibitem[Alemany et~al.(2019)Alemany, Beltran, Perez, and Ganzfried]{alemany2019predicting}
Sheila Alemany, Jonathan Beltran, Adrian Perez, and Sam Ganzfried.
\newblock Predicting hurricane trajectories using a recurrent neural network.
\newblock In \emph{Proceedings of the AAAI Conference on Artificial Intelligence}, 2019.

\bibitem[Bi et~al.(2023)Bi, Xie, Zhang, Chen, Gu, and Tian]{bi2023accurate}
Kaifeng Bi, Lingxi Xie, Hengheng Zhang, Xin Chen, Xiaotao Gu, and Qi~Tian.
\newblock Accurate medium-range global weather forecasting with 3d neural networks.
\newblock \emph{Nature}, pp.\  1--6, 2023.

\bibitem[Brown et~al.(2012)Brown, Milton, Cullen, Golding, Mitchell, and Shelly]{brown2012unified}
Andrew Brown, Sean Milton, Mike Cullen, Brian Golding, John Mitchell, and Ann Shelly.
\newblock Unified modeling and prediction of weather and climate: A 25-year journey.
\newblock \emph{Bulletin of the American Meteorological Society}, 93\penalty0 (12):\penalty0 1865--1877, 2012.

\bibitem[Chen et~al.(2023)Chen, Li, Yang, and Zhang]{chen2023evaluation}
G.~Chen, T.~Li, M.~Yang, and X.~Zhang.
\newblock Evaluation of western north pacific typhoon track forecasts in global and regional models during the 2021 typhoon season.
\newblock \emph{Atmosphere}, 14\penalty0 (3):\penalty0 499, 2023.

\bibitem[Chen et~al.(2020)Chen, Zhang, and Wang]{chen2020machine}
Rui Chen, Weimin Zhang, and Xiang Wang.
\newblock Machine learning in tropical cyclone forecast modeling: A review.
\newblock \emph{Atmosphere}, 11\penalty0 (7):\penalty0 676, 2020.

\bibitem[Chen \& Zhang(2019)Chen and Zhang]{chen2019development}
Xingchao Chen and Fuqing Zhang.
\newblock Development of a convection-permitting air-sea-coupled ensemble data assimilation system for tropical cyclone prediction.
\newblock \emph{Journal of Advances in Modeling Earth Systems}, 11\penalty0 (11):\penalty0 3474--3496, 2019.

\bibitem[Dee et~al.(2011)Dee, Uppala, Simmons, Berrisford, Poli, Kobayashi, Andrae, Balmaseda, Balsamo, Bauer, et~al.]{dee2011era}
Dick~P Dee, S~M Uppala, Adrian~J Simmons, Paul Berrisford, Paul Poli, Shinya Kobayashi, U~Andrae, MA~Balmaseda, G~Balsamo, d~P Bauer, et~al.
\newblock The era-interim reanalysis: Configuration and performance of the data assimilation system.
\newblock \emph{Quarterly Journal of the royal meteorological society}, 137\penalty0 (656):\penalty0 553--597, 2011.

\bibitem[Dendorfer et~al.(2021)Dendorfer, Elflein, and Leal-Taix{\'e}]{dendorfer2021mg}
Patrick Dendorfer, Sven Elflein, and Laura Leal-Taix{\'e}.
\newblock Mg-gan: A multi-generator model preventing out-of-distribution samples in pedestrian trajectory prediction.
\newblock In \emph{Proceedings of the IEEE/CVF International Conference on Computer Vision}, 2021.

\bibitem[ECMWF(2021)]{ecmwf2021tropical}
ECMWF.
\newblock Tropical cyclone activities at ecmwf.
\newblock \url{https://www.ecmwf.int/sites/default/files/elibrary/2021/20228-tropical-cyclone-activities-ecmwf.pdf}, 2021.

\bibitem[Elfwing et~al.(2018)Elfwing, Uchibe, and Doya]{elfwing2018sigmoid}
Stefan Elfwing, Eiji Uchibe, and Kenji Doya.
\newblock Sigmoid-weighted linear units for neural network function approximation in reinforcement learning.
\newblock \emph{Neural networks}, 107:\penalty0 3--11, 2018.

\bibitem[Emanuel(2007)]{emanuel2007environmental}
Kerry Emanuel.
\newblock Environmental factors affecting tropical cyclone power dissipation.
\newblock \emph{Journal of Climate}, 20\penalty0 (22):\penalty0 5497--5509, 2007.

\bibitem[Espeholt et~al.(2022)Espeholt, Agrawal, S{\o}nderby, Kumar, Heek, Bromberg, Gazen, Carver, Andrychowicz, Hickey, et~al.]{espeholt2022deep}
Lasse Espeholt, Shreya Agrawal, Casper S{\o}nderby, Manoj Kumar, Jonathan Heek, Carla Bromberg, Cenk Gazen, Rob Carver, Marcin Andrychowicz, Jason Hickey, et~al.
\newblock Deep learning for twelve hour precipitation forecasts.
\newblock \emph{Nature communications}, 13\penalty0 (1):\penalty0 1--10, 2022.

\bibitem[Giffard-Roisin et~al.(2020)Giffard-Roisin, Yang, Charpiat, Kumler~Bonfanti, K{\'e}gl, and Monteleoni]{giffard2020tropical}
Sophie Giffard-Roisin, Mo~Yang, Guillaume Charpiat, Christina Kumler~Bonfanti, Bal{\'a}zs K{\'e}gl, and Claire Monteleoni.
\newblock Tropical cyclone track forecasting using fused deep learning from aligned reanalysis data.
\newblock \emph{Frontiers in big Data}, pp.\ ~1, 2020.

\bibitem[Golchoubian et~al.(2023)Golchoubian, Ghafurian, Dautenhahn, and Azad]{golchoubian2023pedestrian}
Mahsa Golchoubian, Moojan Ghafurian, Kerstin Dautenhahn, and Nasser~Lashgarian Azad.
\newblock Pedestrian trajectory prediction in pedestrian-vehicle mixed environments: A systematic review.
\newblock \emph{IEEE Transactions on Intelligent Transportation Systems}, 2023.

\bibitem[Gu et~al.(2022)Gu, Chen, Li, Lin, Rao, Zhou, and Lu]{gu2022stochastic}
Tianpei Gu, Guangyi Chen, Junlong Li, Chunze Lin, Yongming Rao, Jie Zhou, and Jiwen Lu.
\newblock Stochastic trajectory prediction via motion indeterminacy diffusion.
\newblock In \emph{Proceedings of the IEEE/CVF Conference on Computer Vision and Pattern Recognition}, 2022.

\bibitem[Gupta et~al.(2018)Gupta, Johnson, Fei-Fei, Savarese, and Alahi]{gupta2018social}
Agrim Gupta, Justin Johnson, Li~Fei-Fei, Silvio Savarese, and Alexandre Alahi.
\newblock Social gan: Socially acceptable trajectories with generative adversarial networks.
\newblock In \emph{Proceedings of the IEEE/CVF Conference on Computer Vision and Pattern Recognition}, 2018.

\bibitem[Huang et~al.(2022)Huang, Bai, Chan, and Zhang]{huang2022mmstn}
Cheng Huang, Cong Bai, Sixian Chan, and Jinglin Zhang.
\newblock Mmstn: A multi-modal spatial-temporal network for tropical cyclone short-term prediction.
\newblock \emph{Geophysical Research Letters}, 49\penalty0 (4):\penalty0 e2021GL096898, 2022.

\bibitem[Huang et~al.(2023)Huang, Bai, Chan, Zhang, and Wu]{huang2023mgtcf}
Cheng Huang, Cong Bai, Sixian Chan, Jinglin Zhang, and YuQuan Wu.
\newblock Mgtcf: Multi-generator tropical cyclone forecasting with heterogeneous meteorological data.
\newblock In \emph{Proceedings of the AAAI Conference on Artificial Intelligence}, 2023.

\bibitem[Huang et~al.(2019)Huang, Bi, Li, Mao, and Wang]{huang2019stgat}
Yingfan Huang, Huikun Bi, Zhaoxin Li, Tianlu Mao, and Zhaoqi Wang.
\newblock Stgat: Modeling spatial-temporal interactions for human trajectory prediction.
\newblock In \emph{Proceedings of the IEEE/CVF International Conference on Computer Vision}, 2019.

\bibitem[IPCC(2023)]{ipcc2023}
IPCC.
\newblock Ar6 synthesis report: Climate change 2023.
\newblock \url{https://www.ipcc.ch/report/ar6/syr/downloads/report/IPCC_AR6_SYR_FullVolume.pdf}, 2023.

\bibitem[Jiang et~al.(2023)Jiang, Cornman, Park, Sapp, Zhou, Anguelov, et~al.]{jiang2023motiondiffuser}
Chiyu Jiang, Andre Cornman, Cheolho Park, Benjamin Sapp, Yin Zhou, Dragomir Anguelov, et~al.
\newblock Motiondiffuser: Controllable multi-agent motion prediction using diffusion.
\newblock In \emph{Proceedings of the IEEE/CVF Conference on Computer Vision and Pattern Recognition}, 2023.

\bibitem[Knapp et~al.(2010)Knapp, Kruk, Levinson, Diamond, and Neumann]{knapp2010international}
Kenneth~R Knapp, Michael~C Kruk, David~H Levinson, Howard~J Diamond, and Charles~J Neumann.
\newblock The international best track archive for climate stewardship (ibtracs) unifying tropical cyclone data.
\newblock \emph{Bulletin of the American Meteorological Society}, 91\penalty0 (3):\penalty0 363--376, 2010.

\bibitem[Lam et~al.(2022)Lam, Sanchez-Gonzalez, Willson, Wirnsberger, Fortunato, Pritzel, Ravuri, Ewalds, Alet, Eaton-Rosen, et~al.]{lam2022graphcast}
Remi Lam, Alvaro Sanchez-Gonzalez, Matthew Willson, Peter Wirnsberger, Meire Fortunato, Alexander Pritzel, Suman Ravuri, Timo Ewalds, Ferran Alet, Zach Eaton-Rosen, et~al.
\newblock Graphcast: Learning skillful medium-range global weather forecasting.
\newblock \emph{arXiv preprint arXiv:2212.12794}, 2022.

\bibitem[Lorenz(1963)]{lorenz1963deterministic}
Edward~N Lorenz.
\newblock Deterministic nonperiodic flow.
\newblock \emph{Journal of atmospheric sciences}, 20\penalty0 (2):\penalty0 130--141, 1963.

\bibitem[Mangalam et~al.(2020)Mangalam, Girase, Agarwal, Lee, Adeli, Malik, and Gaidon]{mangalam2020not}
Karttikeya Mangalam, Harshayu Girase, Shreyas Agarwal, Kuan-Hui Lee, Ehsan Adeli, Jitendra Malik, and Adrien Gaidon.
\newblock It is not the journey but the destination: Endpoint conditioned trajectory prediction.
\newblock In \emph{Proceedings of European Conference on Computer Vision}. Springer, 2020.

\bibitem[Mavrogiannis et~al.(2023)Mavrogiannis, Baldini, Wang, Zhao, Trautman, Steinfeld, and Oh]{mavrogiannis2023core}
Christoforos Mavrogiannis, Francesca Baldini, Allan Wang, Dapeng Zhao, Pete Trautman, Aaron Steinfeld, and Jean Oh.
\newblock Core challenges of social robot navigation: A survey.
\newblock \emph{ACM Transactions on Human-Robot Interaction}, 12\penalty0 (3):\penalty0 1--39, 2023.

\bibitem[Mohamed et~al.(2020)Mohamed, Qian, Elhoseiny, and Claudel]{mohamed2020social}
Abduallah Mohamed, Kun Qian, Mohamed Elhoseiny, and Christian Claudel.
\newblock Social-stgcnn: A social spatio-temporal graph convolutional neural network for human trajectory prediction.
\newblock In \emph{Proceedings of the IEEE/CVF Conference on Computer Vision and Pattern Recognition}, 2020.

\bibitem[Nguyen et~al.(2023)Nguyen, Brandstetter, Kapoor, Gupta, and Grover]{nguyen2023climax}
Tung Nguyen, Johannes Brandstetter, Ashish Kapoor, Jayesh~K Gupta, and Aditya Grover.
\newblock Climax: A foundation model for weather and climate.
\newblock In \emph{International Conference on Learning Representations}, 2023.

\bibitem[Pathak et~al.(2022)Pathak, Subramanian, Harrington, Raja, Chattopadhyay, Mardani, Kurth, Hall, Li, Azizzadenesheli, et~al.]{pathak2022fourcastnet}
Jaideep Pathak, Shashank Subramanian, Peter Harrington, Sanjeev Raja, Ashesh Chattopadhyay, Morteza Mardani, Thorsten Kurth, David Hall, Zongyi Li, Kamyar Azizzadenesheli, et~al.
\newblock Fourcastnet: A global data-driven high-resolution weather model using adaptive fourier neural operators.
\newblock \emph{arXiv preprint arXiv:2202.11214}, 2022.

\bibitem[R{\"u}ttgers et~al.(2019)R{\"u}ttgers, Lee, Jeon, and You]{ruttgers2019prediction}
Mario R{\"u}ttgers, Sangseung Lee, Soohwan Jeon, and Donghyun You.
\newblock Prediction of a typhoon track using a generative adversarial network and satellite images.
\newblock \emph{Scientific reports}, 9\penalty0 (1):\penalty0 6057, 2019.

\bibitem[Salzmann et~al.(2020)Salzmann, Ivanovic, Chakravarty, and Pavone]{salzmann2020trajectron++}
Tim Salzmann, Boris Ivanovic, Punarjay Chakravarty, and Marco Pavone.
\newblock Trajectron++: Dynamically-feasible trajectory forecasting with heterogeneous data.
\newblock In \emph{Proceedings of European Conference on Computer Vision}. Springer, 2020.

\bibitem[Shi et~al.(2021)Shi, Wang, Long, Zhou, Zhou, Niu, and Hua]{shi2021sgcn}
Liushuai Shi, Le~Wang, Chengjiang Long, Sanping Zhou, Mo~Zhou, Zhenxing Niu, and Gang Hua.
\newblock Sgcn: Sparse graph convolution network for pedestrian trajectory prediction.
\newblock In \emph{Proceedings of the IEEE/CVF Conference on Computer Vision and Pattern Recognition}, 2021.

\bibitem[Stull(2011)]{stull2011meteorology}
Roland Stull.
\newblock \emph{Meteorology for Scientists and Engineers}.
\newblock Univ. of British Columbia, 3 edition, 2011.
\newblock ISBN 978-0-88865-178-5.

\bibitem[Walsh et~al.(2016)Walsh, McBride, Klotzbach, Balachandran, Camargo, Holland, Knutson, Kossin, Lee, Sobel, et~al.]{walsh2016tropical}
Kevin~JE Walsh, John~L McBride, Philip~J Klotzbach, Sethurathinam Balachandran, Suzana~J Camargo, Greg Holland, Thomas~R Knutson, James~P Kossin, Tsz-cheung Lee, Adam Sobel, et~al.
\newblock Tropical cyclones and climate change.
\newblock \emph{Wiley Interdisciplinary Reviews: Climate Change}, 7\penalty0 (1):\penalty0 65--89, 2016.

\bibitem[Wang et~al.(2015)Wang, Rao, Tan, and Sch{\"o}nemann]{wang2015statistical}
Yuqing Wang, Yunjie Rao, Zhe-Min Tan, and Daria Sch{\"o}nemann.
\newblock A statistical analysis of the effects of vertical wind shear on tropical cyclone intensity change over the western north pacific.
\newblock \emph{Monthly Weather Review}, 143\penalty0 (9):\penalty0 3434--3453, 2015.

\bibitem[Zhou et~al.(2022)Zhou, Zhu, Hou, Fu, Li, Guan, Sinsky, Kolczynski, Xue, Luo, et~al.]{zhou2022development}
Xiaqiong Zhou, Yuejian Zhu, Dingchen Hou, Bing Fu, Wei Li, Hong Guan, Eric Sinsky, Walter Kolczynski, Xianwu Xue, Yan Luo, et~al.
\newblock The development of the ncep global ensemble forecast system version 12.
\newblock \emph{Weather and Forecasting}, 37\penalty0 (6):\penalty0 1069--1084, 2022.

\end{thebibliography}
\bibliographystyle{iclr2024_conference}
\newpage

\end{document}